\documentclass[runningheads]{llncs}
\usepackage[T1]{fontenc}
\usepackage{graphicx}
\usepackage{subcaption}
\usepackage{multirow}
\usepackage{booktabs}
\usepackage{amsmath}
\usepackage{xcolor}
\usepackage{tabularx}
\usepackage{amssymb}
\usepackage{orcidlink}
\newcommand{\datasetacron}{COCO-PFS}
\newcommand{\datasetname}{COCO Person FaceSwap}

\newcommand\ourarchitectureacronym{Id-CLIP}
\newcommand\ourarchitecturename{Identity-aware CLIP}

\begin{document}
\title{Towards Identity-Aware Cross-Modal Retrieval: \\a Dataset and a Baseline}
\titlerunning{Towards Identity-Aware Cross-Modal Retrieval}
\author{Nicola Messina\thanks{Corresponding Author. Email: \email{nicola.messina@isti.cnr.it}}\inst{1}\orcidlink{0000-0003-3011-2487} \and
Lucia Vadicamo\inst{1}\orcidlink{0000-0001-7182-7038} \and
Leo Maltese \inst{2}\and 
Claudio Gennaro \inst{1}\orcidlink{0000-0002-3715-149X}}

\authorrunning{N. Messina et al.}
\institute{Institute of Information Science and Technologies, CNR, Pisa, Italy\\ \email{\{nicola.messina, lucia.vadicamo,claudio.gennaro\}@isti.cnr.it} \\
\and  University of Pisa, Pisa, Italy - \email{l.maltese1@studenti.unipi.it}}
%\institute{
%Springer Heidelberg, Tiergartenstr. 17, 69121 Heidelberg, Germany
%\and
%\email{\{abc,lncs\}@uni-%heidelberg.de}}
%
\maketitle              % typeset the header of the contribution
\begin{abstract}
%The abstract should briefly summarize the contents of the paper in

Recent advancements in deep learning have significantly enhanced content-based retrieval methods, notably through models like CLIP that map images and texts into a shared embedding space.
However, these methods often struggle with domain-specific entities and long-tail concepts absent from their training data, particularly in identifying specific individuals. In this paper, we explore the task of \textit{identity-aware cross-modal retrieval}, which aims to retrieve images of persons in specific contexts based on natural language queries. 
This task is critical in various scenarios, such as for searching and browsing personalized video collections or large audio-visual archives maintained by national broadcasters. 
We introduce a novel dataset, \datasetname{} (\datasetacron{}), derived from the widely used COCO dataset and enriched with deepfake-generated faces from VGGFace2. This dataset addresses the lack of large-scale datasets needed for training and evaluating models for this task. 
Our experiments assess the performance of different CLIP variations repurposed for this task, including our architecture, \ourarchitecturename{} (\ourarchitectureacronym{}), which achieves competitive retrieval performance through targeted fine-tuning.
Our contributions lay the groundwork for more robust cross-modal retrieval systems capable of recognizing long-tail identities and contextual nuances. Data and code are available at \url{https://github.com/mesnico/IdCLIP}.

\end{abstract}
\section{Introduction}
In modern content-based retrieval literature, deep learning techniques that map images and texts into the same common space -- like CLIP \cite{clip_2021} -- are gaining increasing interest and have been used in many critical downstream multimodal scenarios. 
These models are particularly effective for cross-modal retrieval, enabling efficient searches for relevant images based on natural language queries using straightforward $k$-nearest neighbor techniques in the forged common space.  Consequently, they are widely used in large-scale image or video retrieval, allowing users to easily explore large visual collections using natural language descriptions \cite{vadicamo2024evaluating}. 
%These kinds of networks find an immediate application in cross-modal retrieval, as they are able to effectively and efficiently find the most relevant images representing a given textual query by simply performing a simple $k$-nearest neighbor search in the forged common space, with the possibility of using all the benefits of approximated nearest neighbor search [] to trade effectiveness for efficiency. 
%For these reasons, these methods are nowadays largely employed in large-scale image or video retrieval scenarios, where users can easily explore large visual collections by easily writing a natural language description of the desired scene \cite{vadicamo2024evaluating}. 

 %Despite their widespread adoption, in many downstream tasks, these methods still present many limitations. One of the key issues 
However, despite their widespread adoption, these methods exhibit significant limitations. 
A major challenge is their inability to handle queries containing domain-specific entities and long-tail concepts that were not present within their training set. 
%In other words, methods like CLIP cannot be easily re-purposed to recognize specific persons, objects, or buildings that were not used as training data -- for example, to return relevant results to the query ``\textit{The young singer Gianni Morandi sings at the Sanremo Festival in Italy, black and white footage}''. 
For instance, CLIP struggles to identify specific individuals, objects, or buildings not included during training, which can hinder performance in scenarios like the query, ``\textit{Gianni Morandi sings and dances at the Sanremo Festival in Italy, black and white footage}''.
This limitation becomes particularly important in personalization scenarios, where users seek to locate specific individuals or places within their video collections, as well as in cultural heritage and audiovisual content preservation applications that require retrieving footage of distinct monuments or historical figures in large archives maintained, for example, by national broadcasters.
%This is a strong limitation appearing in many use cases. For example, in the personalization scenario, such an ability could enable finding specific persons or places that show up in a personal video collection. 
%In the context of cultural heritage and audiovisual content preservation, such ability would be the key to retrieving video shots of specific monuments and historical figures in large archives maintained, for example, by national televisions.
%%

Among all the possible long-tail concepts, the ability to recognize people's identities is one of the most critical, with great applicability in many of the above-cited scenarios.
% In this work, we propose a novel dataset, a benchmark, and a baseline as combined steps toward a better understanding of multimodal models' capabilities in the presence of different person identities. Nevertheless, the same generic pipeline can be adapted to data containing other kinds of long-tail concepts like specific buildings, places, paintings, etc.
%
%Some works [] recently started exploring such important limitations of cross-modal retrieval models and proposed some clever solutions. Nevertheless, there are two main problems arising. 
Although some studies \cite{cohen2022my,korbar2022personalised} have already taken some steps in this direction, the main obstacle remains the lack of sufficient high-quality data to fine-tune a multimodal model and make it more discriminative with respect to identity features. Additionally, evaluating these models poses its own challenges, as it is not enough to merely retrieve images of a particular person; the context specified in the query must also be taken into account.
%Another underestimated problem is the careful evaluation of such models, given that, in this scenario, it is important to not only retrieve images containing a specific person but also to retrieve them in the context requested by the query text. 
In the above-mentioned example, we would like not just to retrieve all the possible images depicting \textit{Gianni Morandi}, but only the ones in which he is \textit{singing and dancing} in black and white footage. 

% A simple high-level figure depicting the use of such a model can be seen in Figure []. Different from the original CLIP, \ourarchitecture{} relies on an existing external database of feature vectors representing different persons' features, which complement the data already present in the CLIP weights with external information.

To tackle these challenges, we propose a novel dataset, \datasetacron{}, derived from COCO \cite{lin2014microsoft} and containing persons whose faces have been replaced with entities from VGGFace2 \cite{cao2018vggface2} employing deepfake generation techniques.
Together with the dataset, we introduce suitable metrics able to monitor both the retrieval of the scene context and the person's identity.
We experiment with different CLIP variations to show the challenges this benchmark poses. We demonstrate that the original CLIP  struggles to differentiate between identities. We also examine CLIP-PAD \cite{korbar2022personalised}, a recent CLIP variation that represents a significant step toward creating a domain-independent identity-aware model, capable of taking both visual features of a person's face and text describing the context as input to generate similarity scores with each dataset image. 
%Specifically, we test the original CLIP to show that it is almost unable to comprehend and distinguish different identities. 
%We then probe a recent CLIP variation -- called CLIP-PAD, introduced in \cite{korbar2022personalised} -- which proposes the first step towards the creation of a domain-independent identity-aware CLIP. This architecture can generalize well to any identity proposed at inference time. It works by taking as input a pair composed of the visual feature of the face of the person we desire to retrieve and the text describing the context, and it outputs a similarity each dataset image has with this input pair. 
On top of this solution, we devise our \ourarchitecturename{} (\ourarchitectureacronym{}), which features an effective, clever fine-tuning strategy for the visual backbone to further boost the results, obtaining a strong baseline for this challenging task.
%new task on this challenging benchmark. 
%also of the that of CLIP that can recognize and retrieve a specific set of person identities. We call this architecture Identity-Aware CLIP 

%One simple solution to this problem would be fine-tuning pre-trained generic CLIP models to handle the entities present in a specific scenario. However, this poses many important issues. Fine-tuning is an expensive operation both from a data perspective -- as relatively high volumes of domain-specific image-text pairs should be collected -- and from efficiency and usability perspectives, given that fine-tuning takes time and may be unfeasible for the final user owning the domain-specific data. %Furthermore, fine-tuning CLIP models has the disadvantage of partially destroying important knowledge already acquired through their large pre-training.

To summarize, the contributions of this paper are as follows:
\begin{itemize}
    \item We introduce a novel dataset, \datasetacron{}, featuring images with faces replaced by a controlled set of public figures.
    %derived from COCO, obtained by selecting only images containing persons and replacing their faces with a controlled dataset of public person faces.
    \item We identify a suitable set of metrics able to assess the system's ability to retrieve both the person's identity and the context.
    \item We run multiple experiments on various CLIP models, also introducing \ourarchitectureacronym{} which utilizes fine-tuning of the visual backbone in order to create a strong baseline and capture insights on this novel challenging task.
\end{itemize}

%The rest of this article is structured as follows. We review related work and dataset in Sec.\ref{sec:related_work}; we present our novel dataset in Sec. \ref{sec:dataset}; we present our study for identy-aware retrieval in Section \ref{sec:method} and experimental evaluation in Section \ref{sec:exp}. We report conclusions in Section \ref{sec:conclusions}.

\section{Related Work}\label{sec:related_work}
%\paragraph{Personalized vision-language models}
In recent years, vision-language models, and especially image-text matching methods, have gained significant attention for their application in cross-modal retrieval scenarios. Many of this %text-image matching 
methods adopt hinge-based triplet ranking loss with hard-negative mining using late-fusion approaches \cite{faghri2018vse++,li2019visual,qu2020context,stefanini2021novel,wen2020learning,messina2022aladin,sarafianos2019adversarial,clip_2021,jia2021scaling} or effective yet less efficient early-fusion methods \cite{beit3,li2021align,li2020oscar,lu2019vilbert,Su2020VL-BERT,zhang2021vinvl}.
%
%The rise of Transformers in vision tasks \cite{dosovitskiy2020vit} has led to early-fusion models that process images and text jointly, treating image-text matching as a binary classification problem . Despite strong performance, these models are often computationally intensive for large-scale inference.
%
Among late-fusion methods, CLIP \cite{clip_2021} has become particularly influential for image-text matching and a strong backbone to solve many downstream vision-language tasks. %\cite{liu2023unleashing} proposes a model to find persons given textual descriptions but ignoring contextual information.
%Despite their effectiveness in generic cross-modal retrieval, all of these methods cannot be employed directly to search for long-tail or out-of-distribution concepts unless largely fine-tuned. 
However, these methods struggle with long-tail or out-of-distribution concepts unless extensively fine-tuned.

Most of the personalized vision-language research has focused on personalized image generation. For instance, \cite{galimage,pang2024cross} employed textual inversion to improve personalized feature,  %, with variations in initialization methods. Instead, 
while \cite{wei2023elite,gal2023encoder} proposed encoder-based methods for efficient and accurate customization. Perfusion \cite{tewel2023key} integrated multiple concepts while preventing overfitting, improving inference speed. Other approaches employ multi-concept extraction from a single image \cite{avrahami2023break}, user history-based prompt rewriting \cite{chen2024tailored}, and fine-tuning-free method \cite{zeng2024jedi}. %, contribute to advancing personalized text-to-image models. %Despite these advances, our work focuses on attending to face identity, offering a different angle from object-specific personalization in prior studies.

Some works also started exploring personalized text-image retrieval. Cohen et al. \cite{cohen2022my} introduced a PerVL setup that leverages few-shot learning to reason about personalized concepts. Differently from our work, they update an already existing generic model, fine-tuning it on a small set of specific concepts. Yeh et al. \cite{yeh2023meta} %tackled personalized video searches by learning 
developed instance-specific embeddings without requiring annotated data, enabling instance retrieval in personal videos through vision-language similarity.
%The work most similar to our idea is 
The work most similar to ours is by \cite{korbar2022personalised}, which proposed a method (\textit{CLIP-PAD}) and a dataset (\textit{Celebrities in Action}) for in-context person retrieval. Nevertheless, the proposed dataset collects only very famous actors that may be already present in the CLIP training dataset. Furthermore, our approach builds upon their method by adding a targeted fine-tuning of the CLIP visual backbone to enhance the fine-grained facial understanding abilities.

\paragraph{Entity-aware datasets}
Most existing computer vision datasets lack named entities in the images or captions, limiting their relevance for identity-aware cross-modal retrieval tasks. While some efforts have leveraged knowledge bases to connect textual names with corresponding entities in multimedia content, they usually involve generic entities, such as \textit{woman}, rather than specific, named individuals \cite{dost2020jointly}. The VELD dataset \cite{zheng2022visual} provides scene graphs linking entities from text to the depicted visual content using DBpedia, but it is not publicly available and covers broader categories such as locations and organizations, which are beyond our scope.
WikiPerson \cite{sun2022visual} offers a dataset of over 48k annotated images with bounding boxes and corresponding Wikidata IDs, but it lacks descriptive captions, limiting its applicability for identity-aware retrieval. COFAR \cite{gatti2022cofar} explores \textit{visual} named entities like business brands and world landmarks, linking them through external knowledge sources,  yet it does not include named entities in its captions, hindering its applicability in our context.  %However, named entities are not explicitly included in the provided captions, hindering its use in our context. 
The recently introduced \textit{This-Is-My} dataset \cite{yeh2023meta} includes annotated video segments with captions for contextualized retrieval, but the available captions are limited to a small query-time dataset split, which only includes 30 instances related to 15 entities.  
Rosasco et al. \cite{rosasco2024concon} proposed \textit{ConCon-Chi}, a benchmark designed to assess the compositional capabilities of models on novel chimeric objects within diverse contexts, revealing limitations in handling new concept-context combinations. However, this is a dataset created in controlled scenarios and has limited variability.

Several other datasets have attempted to link textual names to entities in visual content. For example, \cite{ramanathan2014linking} presented a dataset with manually annotated names and captions in TV episode videos, which would have been relevant to our task but not publicly available. Similarly, \cite{korbar2022personalised} used  \textit{Celebrities in Places} and \textit{Celebrities in Action} datasets, containing images and videos of celebrities in various contexts. While highly relevant to our task, these datasets are currently unavailable for download due to privacy concerns.

%Therefore, to address the lack of a large-scale dataset with both images and captions explicitly mentioning person identities, we propose \datasetacron{} dataset specifically designed to support identity-aware cross-modal retrieval and facilitate model development and evaluation.

\section{\datasetname{} (\datasetacron{}) Dataset} \label{sec:dataset}
Current widely-used datasets for text-to-image retrieval \cite{lin2014microsoft,young2014image,sharma2018conceptual} generally contain images with generic descriptions, which are not suitable for identity-aware tasks. These datasets often feature random web-scraped images, making it difficult to find multiple images of the same person, and rarely include names in captions due to privacy concerns. 
%are composed of images accompanied by generic text describing them. These datasets do not immediately fit our needs. In fact, as the images are usually randomly scraped from the web, it is really difficult to find more than one image depicting the same person. Furthermore, the name of the person rarely appears in the textual caption, and this is also due to the privacy concerns public datasets should adhere to. 
These elements are of critical importance for our purpose, which is to understand to which extent state-of-the-art text-to-image retrieval models are able to find the same identity in different contexts.

%Despite the inadequacy of current text-image datasets for our novel identity-aware cross-modal retrieval task, these datasets can be used as a starting point for constructing identity-aware text-image datasets.
%%COCO comes with many images of people together with contextual descriptions. However, persons represented in COCO are generic, and there are no repetitions of the same person instance across different pictures. This is critical for benchmarking the ability of state-of-the-art text-to-image retrieval methods to find the same persons in specific contexts. 
To address this gap, we developed a novel dataset, \datasetname{}, which transform an existing text-image dataset into an identity-aware benchmark.
%Therefore, we devise an automatic, reliable pipeline that 
Our approach includes an automatic pipeline that: (1) selects from the provided dataset only the images containing at least one person, (2) swaps faces with identities from a public face database using deepfake techniques, and (3) enriches captions to include the substituted identity’s name. 
This section describes in detail the stages of the pipeline we utilized to create our benchmark %,  referred to as \datasetname{} dataset, 
that better suits the evaluation of identity-aware cross-modal retrieval.

%%(\textit{image pre-selection})substitutes their faces with the ones from a well-known face database using deepfake approaches (\textit{face-swapping}). The captions are subsequently modified to include the name of the identity substituted in the image (\textit{caption enrichment}).
%
%%In this work, we use the widely employed COCO [] text-image dataset, although many other similar datasets could be used, like LAION [] or WIT [].  
%%augment person images from the original COCO dataset to create a benchmark that better suits our personalized retrieval needs, referred to as \datasetname{} dataset.

%Specifically, the selection of person images and their subsequent face-swapping with known identities generates two datasets that will be important for our study. They are described in the following sections.

% \begin{itemize}
%     \item \textbf{COCO Person Dataset (COCO-P)}: In this version, we produce a filtered version of the COCO dataset, specifically selecting only the images containing a single person or a single face.
    
%     \item \textbf{COCO Person FaceSwap Dataset (COCO-PFS)}: We build on top of the COCO Person Dataset and utilize the Face-swap tool [] in conjunction with the VGGFace2 dataset \cite{vggface} to swap the faces using the ones from the VGGFace2 dataset. This modification allows us to construct a dataset where the same entity appears in different images while also featuring images that depict different entities in the same scenes.
% \end{itemize}

\subsubsection{Image Pre-Selection}
The first step involves selecting images containing individuals, as our focus is solely on retrieving scenes with identifiable persons. %persons, as we are not interested in retrieving scenes with no person identities appearing.
% Given that we are interested in scaling this approach to obtain large-scale datasets for this task in a scripted pipeline, we realized that it is actually difficult to handle more than one person per image. In fact, in the case of two or more persons in the image, (i) it is difficult to automatically propose a modification to the caption to refer to the different identities independently, and (ii) it is difficult to balance the dataset so that the same set of identities appear a consistent number of times in different contexts.  
We selected images from COCO where only a single person appears, given that it is difficult to automatically modify captions to refer to different identities and to ensure a balanced representation of identities across various contexts. To this scope, we first detected faces in the images using MTCNN \cite{zhang2016joint} pretrained on VGGFace2 \cite{cao2018vggface2} and CASIA-Webface \cite{yi2014learning}.  %\textit{FaceNet-Pytorch} library as a face detector\footnote{\url{https://github.com/timesler/facenet-pytorch}}. 
We then filtered out images with multiple detected faces and cross-verified the results using COCO’s annotations to minimize false positives.
%, as a cross-check, we utilized the dataset labels for that image to confirm that there is actually a person in the image -- to drastically diminish the number of false positive detections. 
For each selected image, we extracted the facial features of the depicted individual using an Inception Resnet (V1), also trained on VGGFace2 and CASIA-Webface. % Notice that we cannot employ any original ground-truth COCO detections to reliably crop the faces, as COCO only contains the \textit{person} while lacks the \textit{face} one. %This would be problematic for extracting identity features that we suppose come only from the face. %The captions are modified directly during either the training or inference phases of the model. 

\begin{figure}[t]
 \centering
  \begin{subfigure}[t]{\textwidth}
     \centering
     {\includegraphics[width=\textwidth,trim={0 1.2cm 0 0},clip]{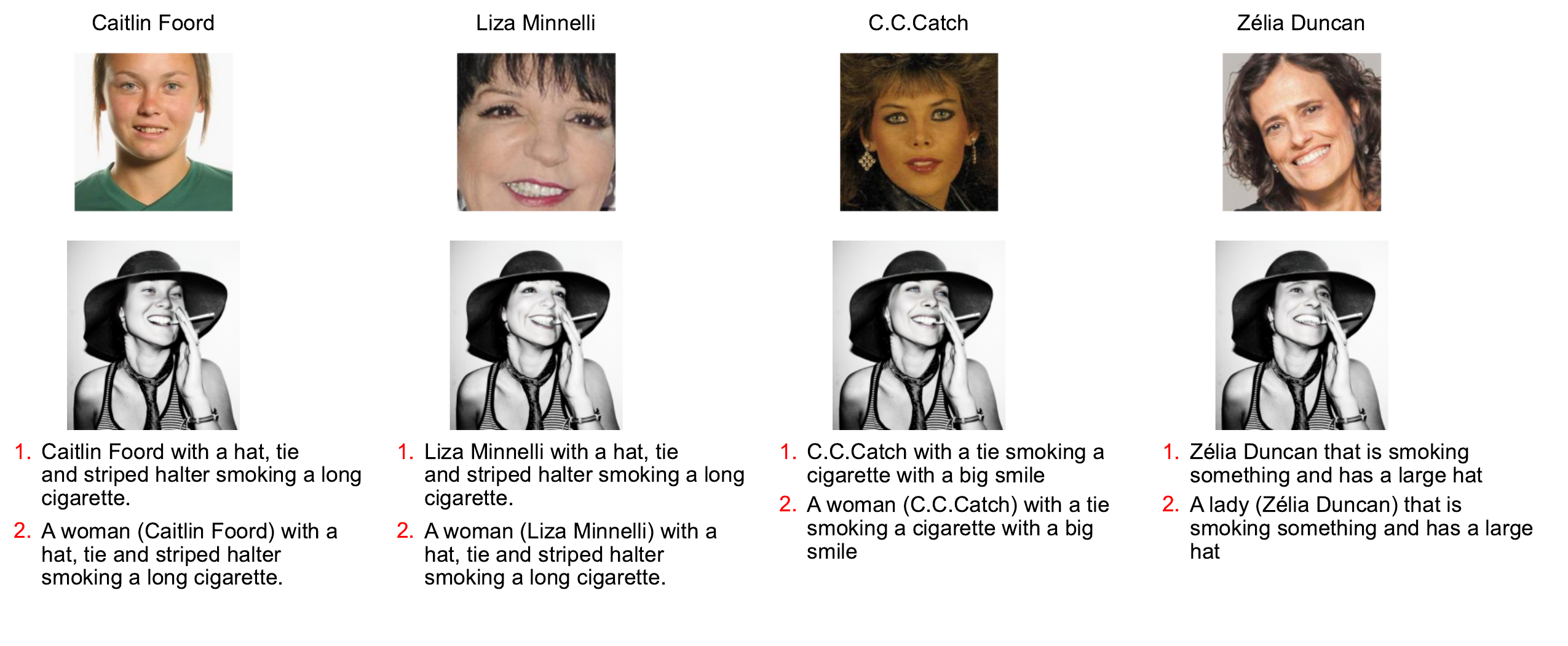}}
     \caption{One context, multiple identities.}
     \label{fig:dataset-single-context}
 \end{subfigure}
 \hfill
 \begin{subfigure}[t]{\textwidth}
     \centering
     \includegraphics[width=\textwidth,trim={0 3.2cm 0 0},clip]{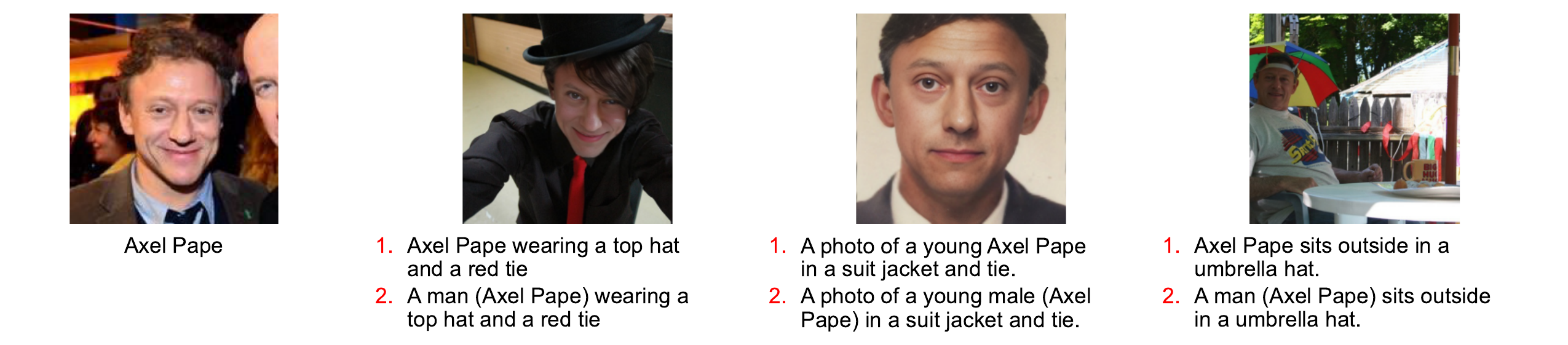}
     \caption{One identity, multiple contexts.}
     \label{fig:dataset-single-entity}
 \end{subfigure}
\caption{Examples from our \datasetacron{} dataset showing images obtained by substituting (a) different identities in the same context or (b) the same identity in different contexts. The captions depicted are obtained through two different templates, starting from one of the five original COCO captions.}
\label{fig:dataset-creation2}
\vspace{-3mm}
\end{figure}

\subsubsection{Gender-Ethnicity Detection and Face-swapping}
%One key stage of our dataset creation pipeline was the actual substitution of the person's faces with the ones from a knowledge base of VIP faces. 
%As the set from which to draw the different identities, we employ the test set of VGGFace2, which contains 500 different identities. 
In this stage, we substitute the faces in COCO images with those from a VIP face database, using the VGGFace2 test set, which includes 500 identities.
To ensure a balanced dataset, we ensure that each identity is present in at least two different contexts, and each image is swapped with multiple faces to depict various people. 
%Similarly, we force the same COCO image to be face-swapped multiple times with different identities so that different persons appear in each context.
%With all these generation procedures set, we employ the 
We used \textit{Roop} tool\footnote{\url{https://github.com/s0md3v/roop}}, an advanced, easy-to-use face-swapping software, to reliably swap the faces. %, creating a balanced arrangement of samples.
%
%In order to allow for realistic swaps, we avoid randomly picking a face from VGGFace2. Instead, we extract gender and ethnicity features from the COCO images using the \textit{DeepFace tool}\footnote{\url{https://github.com/serengil/deepface}} to find a suitable face in VGGFace2 that matches the given anatomical characteristics of the original person.
To make these swaps realistic, we don’t select faces randomly. Instead, we use \textit{DeepFace tool}\footnote{\url{https://github.com/serengil/deepface}} -- which implements advanced face-analysis models -- to detect the gender and ethnicity of the original faces in COCO, selecting a suitable matching face from VGGFace2 for the swap.
Before the actual face-swapping phase, we keep only the images in which the individuals' faces are large in proportion to the image size. This filtering step is crucial to ensure the optimal performance of tools such as Deepface and Roop, particularly when dealing with face swapping. In fact, these tools tend to produce superior results when the facial features are prominently visible, and the face occupies a significant portion of the image.

\subsubsection{Caption Enrichment}
%We transformed the original generic COCO descriptions into a set of potential realistic query captions through a simple template-based approach. 
%Specifically, for each image $I_i$, let $N$ be the name of the person depicted in the image (i.e., the identity whose face was swapped), and let $C_{i1}, \dots, C_{i5}$ be the original COCO captions associated with the image prior to the face swap.
%
%Specifically, we first detect the generic noun in the caption $C_i$ referring to the person in the image -- e.g., \textit{person}, \textit{man}, \textit{woman}, \textit{kid}, etc.
%
%%Then, we derive a set of captions from a given set of templates: $\mathcal{C}^\text{ens}_i = \{\text{template}_j(C_i)\}_{j=1}^k$ -- where $k$ is the number of prepared templates. For example, assuming a set of three different templates, the caption \textit{``A kid is running at the park''} may become: (i) "\textit{[ENTITY] is running at the park}", (ii) "\textit{An image with [ENTITY]. The kid is running at the park}"; (iii) "\textit{The famous kid [ENTITY] is running at the park}".
%%The word \textit{[ENTITY]} is a placeholder for the actual identifier of the person, which could be the actual name or a special token carrying visual information about the face, as will be detailed in Section \label{sec:model}.
%
To create more personalized and realistic captions, we transform COCO's generic descriptions using a template-based approach.
For each image $I_i$, where $N$ is the swapped person's name, we modify the five original COCO captions ($C_{i1}, \dots, C_{i5}$) by replacing generic noun referring to the person in the image (e.g., \textit{person}, \textit{man}, \textit{woman}, \textit{kid}, etc.) with the specific name $N$ using $k$ templates, $\{\mathbf{T}^t\}_{t=1}^{k}$. As a results, we generated $k$ alternative versions of each caption, $C_{ij}^t = \mathbf{T}^t(C_{ij}, N)$, for  $t= 1, \dots, k$.

%For each caption $C_{ij}$, we first detect the generic noun referring to the person in the image (e.g., \textit{person}, \textit{man}, \textit{woman}, \textit{kid}, etc.). Then, we use a set of $k$ templates, $\{\mathbf{T}^t\}_{t=1}^{k}$, to modify the caption into $k$ alternative versions, $C_{ij}^t = \mathbf{T}^t(C_{ij}, N)$, for  $t= 1, \dots, k$.

For example, assuming a set of three different templates, the original caption \textit{``A person is running at the park''} and the name $N$ = \textit{``Gianni Morandi''} can be transformed into: (i) ``\textit{Gianni Morandi is running at the park}'', (ii) ``\textit{An image with Gianni Morandi. The person is running at the park}''; (iii) ``\textit{The famous person Gianni Morandi is running at the park}''. 
%Given that COCO comes with five human written captions for each image, and we use $k$ templates for each caption, we have $5\cdot k$ captions for each image. 
Our dataset has been constructed using $k=11$ different templates for each caption. For ease of notation, we flatten the indexes $j$ and $t$ to obtain again $C_{ij}$, where $j$ this time spans from $1$ to $5\cdot k$.
%The word \textit{[ENTITY]} is a placeholder for the actual identifier of the person, which could be the actual name or a special token carrying visual information about the face, as will be detailed in Section \label{sec:model}.

We also release the captions with a [ENTITY] placeholder on behalf of the actual name so that methods can directly employ this information without having to rely on external NER modules for re-extracting the name.

\subsubsection{Dataset Overview}
We applied the proposed pipeline to the training, validation, and test sets of the COCO dataset. The resulting training set contains 49,957 images with 500 distinct entities, resulting in 2,747,635 captions. The validation and test sets each contain 1,760 images, with 466 and 467 entities, respectively, and 96,800 captions per set. 
We report an example in Figure \ref{fig:dataset-creation2}.

\section{Identity-aware Cross-modal Retrieval} \label{sec:method}
\textit{Identity-aware cross-modal retrieval} extends the classic {text-to-image retrieval} task, where the goal is to find the most relevant images in a potentially large database given a natural language query. In our scenario, textual queries are enriched with person names, and the task is to retrieve all images featuring the specified person in the described context.

% Assuming we have a method capable of processing these natural language queries, we propose to evaluate performance on two distinct sub-tasks. The first sub-task, \textit{context-entity retrieval}, focuses on retrieving images that contain both the requested person and the specified context. Differently,  the second sub-task, \textit{entity-only retrieval}, aims to find all images depicting a given entity, regardless of the specific context in which they appear.

%To evaluate performance on these two sub-tasks, 
%More specifically, we 
 To train or evaluate an identity-aware model, we assume the availability of a \textit{dataset} $\mathcal{D} = \{\left(I_i, \{C_{ij}\}_{j}\right ), i=1\dots, n\}$, which consists of $n$ images $I_i$, each associated with one or more descriptive natural language captions $\{C_{ij}\}$. Additionally, we consider a \textit{face gallery} $\mathcal{G} = \{(F_l, N_l)\}_{l=1}^m$ containing $m$ identities.  Each person in the gallery is represented by an image of their face $F_l$ and their name $N_l$  (e.g., \textit{Mario Rossi}). 
Each image $I_i$ should contain individuals from $\mathcal{G}$ in various contexts, with the corresponding captions $C_{ij}$ explicitly mentioning the person’s name (e.g., \textit{Mario Rossi is riding a bike down the city streets}). This is exactly how our \datasetacron{} dataset is designed. The face gallery -- VGGFace2 in our case -- serves as an external knowledge base, allowing us to retrieve a face $F_l$ given a person’s name $N_l$.
%where given the name of a person $N_l$, the corresponding image of their face $F_l$ can be retrieved. 
In this setting,  given any of the descriptions $C_{ij}$ as a query, the identity-aware retrieval objective is to rank all $n$ images of the dataset by decreasing relevance, ensuring that  $I_{i}$ -- the \textit{correct} image containing the person mentioned in $C_{ij}$ -- is ranked as high as possible.

\begin{figure}[t]
 \centering
 \begin{subfigure}[t]{.43\textwidth}
     \centering
     {\includegraphics[width=\textwidth,page=3]{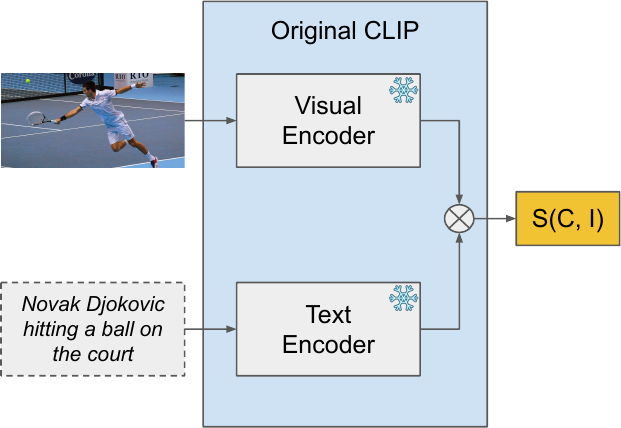}}
     \caption{Original CLIP.}
     \label{fig:original-clip}
 \end{subfigure}
 \hfill \,\,
 \begin{subfigure}[t]{.52\textwidth}
     \centering
     {\includegraphics[width=\textwidth,page=4]{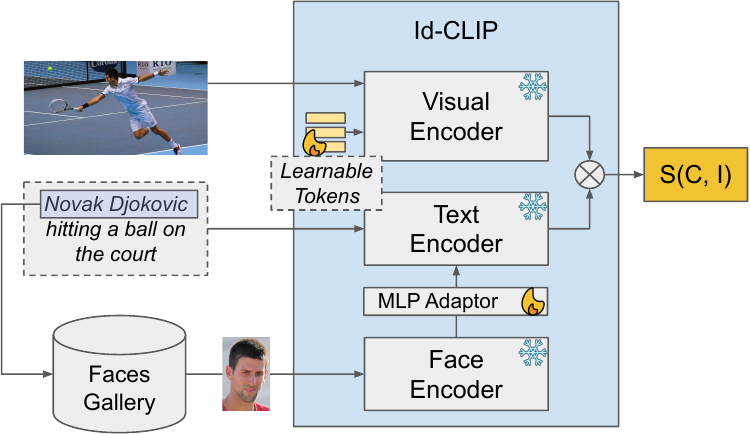}}
     \caption{Identity-Aware CLIP (Id-CLIP).}
     \label{fig:id-clip}
 \end{subfigure}
\caption{(a). The original CLIP model. In order to digest entity names, should be properly finetuned and specialized to a given gallery $\mathcal{G}$. (b) \ourarchitectureacronym{} solves the problem by factoring out the entity from the caption. Through $\mathcal{G}$, the name is used to lookup the face, which becomes an additional input to the system. The learnable tokens in input to the Visual Encoder implement \textit{visual prompt tuning}.}
\label{fig:dataset-creation}
\vspace{-3mm}
\end{figure}

A straightforward solution to this task involves training a text-to-image retrieval model on a dataset containing all identities from $\mathcal{G}$.
%The presented task could ideally be solved by a text-to-image retrieval model trained on a dataset containing all possible persons from the gallery $\mathcal{G}$.
Once trained, the model would contain all the entity-specific knowledge in its weights (Figure \ref{fig:original-clip}). However, this would require re-training the model for each gallery to transfer knowledge of specific person identities into the model weights. While feasible, this is impractical for real-world applications where users may want to fine-tune the system with their own galleries, such as smartphone photo collections.

A much more application-friendly formulation of this problem would use a single general-purpose trained model that adapts, at inference time, to any user-provided gallery of identities $\mathcal{G}$. % provided by the user at inference time.
The idea is to leverage the general knowledge that cross-modal models like CLIP \cite{clip_2021} have about the world -- which helps in understanding generic image-text correspondences -- together with the domain-specific knowledge present in the user-provided gallery $\mathcal{G}$ -- necessary to distinguish between different identities.
%The key idea to achieve this is, at training time,  to 
%
To achieve this, during training, we remove the person's name from the query $C_{ij}$, substituting it with its face instead. This can be done by extracting the person's name $N_{i}$ from $C_{ij}$ using off-the-shelf named entities recognition methods and using it to look up the corresponding face $F_{i}$ in $\mathcal{G}$. 
This is exactly the idea behind \ourarchitectureacronym{}, presented in the next section and summarized in Figure \ref{fig:id-clip}. 
% Within this new framework model, the query becomes multi-modal, as the retrieval model processes a compound query consisting of a possibly rewritten version $\tilde{C}_{ij}$ of the original textual query -- for example, anonymized by converting the person name like \textit{Gianni Morandi} into a generic \textit{person} to better fit the textual encoder training domain -- and the person's face $F_i$. This reformulation effectively prevents multi-modal models from learning an association between the name of the person and the features of the face, given that, in this setup, the model is never presented with the actual person's name during training.  This approach is used to train our \ourarchitectureacronym{} model, which is described in the next section.

\subsection{\ourarchitecturename{} (\ourarchitectureacronym{})}
\label{sec:model}

In this work, we adapt existing CLIP-based models to search for relevant images given a \textit{compound multimodal query} $(\tilde{C}, F)$, where $\tilde{C}$ is an anonymized caption and $F$ represents an image of a person's face. We call this adaptation \textit{\ourarchitecturename{}(\ourarchitectureacronym{})}, which extends the CLIP-PAD architecture from \cite{korbar2022personalised} by incorporating a \textit{visual prompt tuning} stage.  This enhancement serves as a targeted, non-invasive fine-tuning strategy for the CLIP vision encoder, allowing the model to better focus on person identities.
%In the following paragraphs, we first recall the CLIP-PAD architecture introduced in [], and then we discuss the addition of \textit{visual prompt tuning} as a targeted non-invasive fine-tuning strategy of the CLIP vision encoder to better focus on person identities.

\subsubsection{Overview}
\ourarchitectureacronym{} is inspired by the work in \cite{korbar2022personalised}, in which the authors produce an anonymized caption $\tilde{C}$ by substituting the person name in $C$ with a special token called [TOK]. This special token is not associated with a learnable embedding, as is the case with normal word tokens. Instead, its embedding $\mathbf{w}^\text{[TOK]}$ %for the [TOK] token 
is obtained as a multi-layer perceptron (MLP) projection of the visual embedding extracted from the person's face $F$ through a state-of-the-art face recognition network (FRN): $\mathbf{w}^\text{[TOK]} = \text{MLP}(\text{FRN}(F))$.
The whole network is then subject to the same CLIP contrastive learning objective used to pull relevant image-text pairs and repel irrelevant ones.
The network is trained using a suitable dataset $\mathcal{D}^\text{train}$ containing images of persons in specific contexts. During training, the CLIP-PAD network only updates the MLP projection network, while the rest of the CLIP weights are kept frozen. 
The MLP network learns to map the facial features into an actual word token that the frozen pre-trained text encoder can process alongside the other caption tokens. 
%The sole critical part that the authors train is the MLP projection network, with all the remaining weights of CLIP frozen. 
%The MLP network should be able to transform the visual feature of the face into an actual word token that can be processed, together with all the other caption tokens, through the frozen pre-trained text encoder. 
The training phase's sole purpose is to learn meaningful semantic mapping between the face feature space and the CLIP token embedding space.
%\ourarchitectureacronym{} differs from [] in two different important ways. First, we introduce visual prompt tuning from the visual side to adapt the CLIP vision encoder to pay more attention to the actual person's face. Second, we employ prompt ensembling at inference time. We report an overview of \ourarchitectureacronym{} in Figure [], and we better detail these contributions in the following paragraphs.
Therefore, we obtain the caption embedding $\mathbf{c}$ by applying the CLIP text encoder to the sequence of tokens containing the special token $\mathbf{w}^\text{[TOK]}$ together with the original sentence tokens $\mathbf{w}_i$: $\mathbf{c} = \text{CLIP}^\text{txt}([\mathbf{w}_1, \mathbf{w}_2, \dots, \mathbf{w}^\text{[TOK]}, \dots, \mathbf{w}_n])$. Notice that $\mathbf{w}^\text{[TOK]}$ is inserted in specific positions inside the caption $\tilde{C}$, as better explained in Section \ref{sec:exp}.

\subsubsection{Visual Prompt Tuning} 
The rationale behind \ourarchitectureacronym{} is that the person's identities are discriminated solely through their visual features, both at the query and image encoding stages. At query encoding time, the model can just leverage learning a suitable MLP projection network to discriminate between two different queries that carry the same context but different identities -- relying on the discriminative ability of the off-the-shelf face recognition network.
From the image dataset encoding perspective, instead, CLIP-PAD assumes that the visual encoder is already able to discriminate between different identities, assuming equal context. Considering that CLIP has not been directly trained to recognize fine-grained facial differences and that this information is not present in the generic image-text training datasets, it is likely that the CLIP visual encoder fails in such a task.
In order to compensate for this potential drawback, we enrich the visual encoder of \ourarchitectureacronym{} with some additional learnable visual tokens $\tilde{\mathbf{v}}_i$, employing the approach presented in \cite{jia2022visual}. These additional visual tokens are purposed to learn the discriminative characteristics of the person's faces. 
Specifically, the final image feature $\mathbf{v}$ is obtained through the application of the CLIP visual encoder to the original image tokens $\{\mathbf{v}_i\}_{i=1}^{wh}$ concatenated to the series of the special learnable prompt tuning tokens $\{\tilde{\mathbf{v}}_i\}_{i=1}^p$: $\mathbf{v} = \text{CLIP}^\text{vis}([\tilde{\mathbf{v}}_1, \dots, \tilde{\mathbf{v}}_p, \mathbf{v}_1, \dots, \mathbf{v}_{wh}])$, where $p$ is the number of prompt tuning tokens (we empirically found that $p=5$ works best in our scenario), and $wh$ is the total number of visual patches of the image.
Notice that, in this setup, the MLP and the added visual tokens are learned together during the training procedure, while the core CLIP weights remain frozen.
%Notice that the CLIP core weights are always kept frozen.

\subsubsection{Objective}
As our image-text contrastive objective, we employ an Info-NCE loss \cite{oord2018representation}, formulated as follows:
\begin{equation} 
\mathcal{L}_{\text{InfoNCE}} = -\frac{1}{B} \sum_{i=1}^{B} \log \frac{\exp(\text{sim}(\mathbf{c}_i, \textbf{v}_i))}{\sum_{j=1}^{B} \exp(\text{sim}(\mathbf{c}_i, \mathbf{v}_j))}, 
\end{equation}
where $\mathbf{c}_i$ and $\mathbf{v}_j$ are the textual and visual features in output from CLIP, as explained in the previous paragraphs, $\text{sim}(\cdot, \cdot)$ is the cosine similarity, and $B$ is the batch size employed during training.

\section{Experiments} \label{sec:exp}

We evaluate the performance of \ourarchitectureacronym{} mainly on the targeted identity-aware cross-modal retrieval task, which we refer to as the \textit{entity-in-context retrieval} -- that focuses on retrieving images that contain both the requested person and the specified context. However, we can probe the same model also on a second related task, which we call \textit{entity-only retrieval}. This secondary task aims at finding all images depicting a given entity, regardless of the specific context in which they appear. This second task is mainly crafted to understand the abilities of the trained model to actually understand the person's features without the potential noise introduced by the context.

\subsubsection{Inference Strategies}
We apply two different inference strategies depending on the task we are tackling, i.e., \textit{entity-in-context} retrieval or \textit{entity-only} retrieval.

Concerning the entity-in-context retrieval subtask, our proposed \datasetacron{} dataset already comes with the anonymized captions $\tilde{C}_{ij}$, which contain the [ENTITY] placeholder instead of the actual name\footnote{In a real scenario, we should use NER techniques to extract the entity from the query. However, in this setup, we already have this information available considering how we built the dataset.}. In the most simple case, [ENTITY] could be substituted with the [TOK] token expected by \ourarchitectureacronym{}. However, other meaningful strategies may be applied, such as replacing [ENTITY] solely with the original text name (e.g., \textit{Gianni Morandi}), or a combination of textual and visual features -- like \textit{[TOK] Gianni Morandi}, or \textit{Gianni Morandi [TOK]}. Other strategies are possible, like placing the visual feature at the beginning of the string and then substituting [ENTITY] to the sole textual name -- e.g., \textit{The famous kid [ENTITY] is running at the park} may become \textit{[TOK]. The famous kid Gianni Morandi is running at the park.}. For entity-in-context retrieval, we consider each caption $\tilde{C}_{ij}$ as a separate query with which the whole image database is searched, and we tested different templates to substitute [ENTITY] with [TOK] and the original person’s name.

%\datasetacron{} comes with a set $\mathcal{C}^\text{ens}_i$ of $k$ captions for each original image caption in COCO, derived with different rules. We could just choose only one of them at inference time. Instead, similarly to the procedure used in CLIP [], we perform inference by computing the average of the text features extracted from the text encoder for each one of the elements in $\mathcal{C}^\text{ens}_i$. In other words, as the query representation, we are using an average of the features obtained with all the different query templates. This is done to effortlessly increase the performance at inference time using a simple prompting strategy. We are only assuming that it is easy to derive multiple phrasings of the user-provided query at inference time, which is the case in this scenario.

Differently, for entity-only retrieval, the sole query is the name of the person to be searched. In this context, we treat each person as a different categorical class. We can solve this task using \ourarchitectureacronym{} by employing [ENTITY] as the sole input to the model. Since we are interacting with a language model that understands natural language sentences, we surround the [ENTITY] tag with some pre-defined templates, like \textit{An image with [ENTITY]} or \textit{The famous [ENTITY]}, performing \textit{prompt ensembling} like in CLIP \cite{clip_2021}. [ENTITY] is then expanded in the same way as for the entity-in-context retrieval, using a combination of [TOK] and the original person's name. %However, in this case -- similarly to the procedure used in CLIP [] -- we perform inference by performing \textit{prompt ensembling}, i.e., considering as the query vector the average of the text features extracted from the text encoder for each one of the template phrases.

\subsection{Metrics}
% We employ different retrieval metrics to capture the most important aspects of identity-aware cross-modal retrieval. Specifically, we are interested in understanding if the model retrieves both context and identities correctly (\textit{context-entity retrieval}), as well as in tracking how well the model is able to retrieve specific entities unregarding the context (\textit{entity-only retrieval}).

% For context-entity retrieval, we use the standard metric commonly employed to evaluate the cross-modal retrieval models, the \textit{recall@k} metric. Specifically, recall@k measures the percentage of text queries that successfully retrieve the correct image containing the appropriate entity in the right context among the first \textit{k} results. In our experiments, we report this metric with $k \in \{1,5,10,50\}$.

% Concerning entity-only retrieval, we instead compute a recall@k metric for the model prompted solely with the entity, without the context. In particular, provided a query containing a given entity, we compute the percentage of items in the first $k$ results that match the specified identity, unregarding the context:

In the experiments, we report the standard metric commonly employed to evaluate the cross-modal retrieval models, the average recall@k over all the tested queries, varying $k \in \{1,5,10,50\}$.
\begin{equation}
\begin{aligned}
     & \text{Recall@k} =\frac{|\text{Relevant instances} \cap \text{Retrieved top-k instances}|}{\text{min}(k, |\text{Relevant instances}|)}            
\end{aligned}
\end{equation}
In the case of  \textit{entity-in-context retrieval}, we have only one relevant image in the dataset for each query text. Therefore, the average recall@k  corresponds to the percentage of text queries that successfully retrieve the correct image containing the appropriate entity in the right context within the top \textit{k} results.

We also assess model performance on the \textit{entity-only retrieval} task, where the query text includes only the entity’s name without any context.  In this case, the recall@k measures the percentage of items in the top-$k$ results that match the specified identity, regardless of the image's context.

We also report $\text{Rsum} = \sum_{k \in \{1, 5, 10, 50\}}$ \text{Recall@}k as an aggregated measure.

\subsection{Implementation Details}
We optimized \ourarchitectureacronym{} using the Adam optimizer. We employed a learning rate of 5e-5. The MLP used to convert face features in a textual token is composed of one hidden layer without bias weights and ReLU non-linearities. The model is trained for a maximum of 10 epochs and validated on the main entities-in-context retrieval task.
During training, as a robust augmentation strategy, we employ the face crops of the face-swapped persons instead of the original ones as $F$. During testing, instead, $F$ is a real face taken from $\mathcal{G}$.
Concerning captions, at training time, a single template is chosen randomly among all the available ones. This can be considered as an additional text augmentation procedure, which enables the network to generalize to different phrasings of the same (context, identity) pair, in turn stabilizing the learning procedure.

\subsection{Results}

For the entity-in-context retrieval, we report the results in Table \ref{tab:context-entity-results}, which is split into two equal sets of experiments: one without Visual Prompt Tuning (VPT) and one with VPT engaged.  %with Visual Prompt Tuning disengaged and engaged, respectively. 
The first and the second lines report the two trivial baselines for this task, which are represented by the original CLIP prompted with the original caption and by the original CLIP prompted with \datasetacron{} captions containing the person names (CLIP N). 

\newcolumntype{L}{>{\centering\arraybackslash}m{1cm}}%
\newcolumntype{M}{>{\arraybackslash}m{4cm}}%
\newcolumntype{C}{>{\centering\arraybackslash}X}
\begin{table}[t]
\caption{Entity-in-context Retrieval Results}
\label{tab:context-entity-results}
\begin{tabularx}{0.95\linewidth}{MLCCCCC}
\toprule
 & & \multicolumn{4}{c}{Recall@k} & \\
 \cmidrule{3-6}
 Model & VPT & k=1 & k=5 & k=10 & k=50 & Rsum \\
\midrule
CLIP (Original) & \multirow[c]{6}{*}{--} & 12.10 & 56.20 & 69.10 & 92.20 & 229.60 \\
 CLIP N & & 26.20 & 58.20 & 70.10 & 91.80 & 246.30 \\
 CLIP-PAD T (All $\mathbf{T}_i$) & & 36.30 & 66.60 & 77.80 & 95.20 & 275.90 \\
 % & conf6 & 34.20 & 58.90 & 68.60 & 85.10 & 246.80 \\
 CLIP-PAD T+N (All $\mathbf{T}_i$) & & 36.40 & 65.00 & 75.80 & 94.40 & 271.50 \\
 CLIP-PAD T ($\mathbf{T}_1$) & & 39.30 & 69.70 & 80.80 & 96.50 & 286.20 \\
 CLIP-PAD T+N ($\mathbf{T}_1$) & & 38.00 & 67.10 & 78.30 & 95.50 & 278.90 \\
\midrule
CLIP (Original) & \multirow[c]{6}{*}{\checkmark} & 12.40 & 57.90 & 70.60 & 93.20 & 234.10 \\
 CLIP N & & 27.90 & 61.70 & 72.90 & 93.00 & 255.50 \\
 Id-CLIP T (All $\mathbf{T}_i$) & & 39.20 & 68.20 & 78.50 & 95.70 & 281.60 \\
 Id-CLIP T+N (All $\mathbf{T}_i$) & & 37.50 & 66.10 & 76.30 & 94.30 & 274.10 \\
 Id-CLIP T ($\mathbf{T}_1$) & & \bfseries 43.20 & \bfseries 71.20 & \bfseries 80.80 & \bfseries 96.70 & \bfseries 291.90 \\
 Id-CLIP T+N ($\mathbf{T}_1$) & & 39.90 & 68.60 & 78.10 & 95.00 & 281.60 \\
%\cline{1-7}
\bottomrule
\end{tabularx}
\end{table}

\begin{table}[t]
\caption{Entity-only Retrieval Results. In configuration (1), [ENTITY] is expanded as ``[NAME] [TOK]'', while in (2), the sentence always starts with ``An image with [TOK].'' and is then followed by the original caption where [ENTITY] is substituted with [NAME]. }
\label{tab:entity-results}
\begin{tabularx}{0.95\linewidth}{MLCCCCC}
\toprule
 & & \multicolumn{4}{c}{Recall@k} & \\
 \cmidrule{3-6}
 Model & VPT & k=1 & k=5 & k=10 & k=50 & Rsum\\
\midrule
 CLIP N & \multirow[c]{4}{*}{--} & 12.80 & 6.00 & 8.40 & 17.50 & 44.80 \\
 CLIP-PAD T & & 15.40 & 11.10 & 13.50 & 28.00 & 68.00\\
 CLIP-PAD T+N (1) & & 20.40 & 11.80 & 13.90 & 28.00 & 74.10\\
 CLIP-PAD T+N (2) & & 21.80 & 11.80 & 14.20 & 28.20 & 75.90\\
 \midrule
CLIP N & \multirow[c]{4}{*}{\checkmark} & 15.40 & 7.30 & 9.20 & 19.10 & 51.00 \\
 Id-CLIP T & & 21.20 & 11.50 & 14.20 & \bfseries 31.90 & 78.90 \\
 Id-CLIP T+N (1) & & 22.60 & 11.60 & 14.60 & 29.70 & 78.60\\
 Id-CLIP T+N (2) & & \bfseries 24.30 & \bfseries 13.70 & \bfseries 16.20 &  31.40 & \bfseries 85.60 \\
\bottomrule
\end{tabularx}
\end{table}

By focusing on the recall@1 metric, we can notice how CLIP is already able to exploit some biases when prompted with the person's names -- most likely, certain names imply specific ethnical groups, which imply certain facial features. We then first re-evaluate CLIP-PAD (which does not have visual prompt tuning active) on our benchmark, considering the most successful inference strategies. The performance improves significantly with CLIP-PAD (All $T_i$), which averages recall across 11 different templates evaluated independently, resulting in an improvement of more than 38\% compared to CLIP N. %which evaluates all the 11 templates independently obtaining an improvement of more than 38\% with respect to CLIP N. 
The two different reported versions, T and T+N, relate to how [ENTITY] is substituted -- with only [TOK] in the first case and with [NAME] [TOK] in the second one, where the name is the person's name. Interestingly, CLIP-PAD seems not to improve as much as the CLIP baselines when presented with original names. In fact, the difference between CLIP-PAD T and CLIP-PAD T+N is quite marginal with respect to the huge improvement observed in the baselines. This may be due to the ethical biases present in the pre-trained CLIP transforming into unuseful noise when the facial feature is injected through the [TOK] token.
We also report CLIP-PAD ($T_1$) -- coming as well in the T and in T+N configurations, which, differently from CLIP-PAD (All $T_i$), only tests the best-performing template, which is \textit{$T_1$= ``[ENTITY] in the image. [CAPTION]''}. We can notice that the best template behaves particularly better than the ``All $T_i$'' configuration, meaning that, on average, some reasonable query behaves particularly worse than the average. Shifting the attention to the results where visual prompt tuning is active, we can appreciate how much our proposed \ourarchitectureacronym{} improves over CLIP-PAD for all the probed inference schemas. In particular, \ourarchitectureacronym{} obtains an improvement on recall@1 of 55\% with respect to CLIP N, with an average increase over CLIP-PAD of 6\%, proving the effectiveness of this targeted fine-tuning strategy for the visual encoder. Notice also that the improvement on the recall@k for higher values for $k$ is limited given that, at a certain point, the model will have retrieved all the pictures of the different identities in a given context\footnote{We remind that, in our dataset, each image has been swapped with multiple identities, and this applies to the test set as well.}, thus in the limit behaving similarly to the original CLIP model. We obtain the best overall results over all the metrics with the \ourarchitectureacronym{} T ($T_1$) model.

Concerning the entity-only retrieval, we report the results in Table \ref{tab:entity-results}.
%Even in this case, we report the results with VPT disabled and enabled, respectively. 
Again, we can immediately notice how the original CLIP model prompted with names (CLIP N) behaves poorly on this subtask. Nevertheless, differently from the results in entity-in-context retrieval, we can notice how the original name plays a more important role, as far as CLIP-PAD is concerned. In fact, the configurations denoted as CLIP-PAD T+N obtain a significant improvement over CLIP-PAD T, meaning that the name biases present in CLIP help in bringing reasonable identities to the top of the ranked list. This effect is actually damped in \ourarchitectureacronym{}, where visual prompt tuning effectively increases the importance of the visual features over the textual biases. For completeness, we report two different [ENTITY] expansion strategies: T+N (1) represents the case in which [ENTITY] is expanded as ``[NAME] [TOK]''; (2) represents the case where the sentence is always started with ``An image with [TOK].'' and then followed by the original caption where [ENTITY] is substituted with only [NAME]. The difference between the two versions is not so pronounced, meaning that the model seems robust to different entity expansion formulations. Even in this case, we reach the best result through \ourarchitectureacronym{}, again proving the importance of VPT. Specifically, the best-performing model results to be Id-CLIP T+N (2), which increases the recall@1 by 57\% with respect to CLIP N and by 11\% with respect to the corresponding configuration of CLIP-PAD.

\section{Conclusions} \label{sec:conclusions}

In this work, we tackled the \textit{identity-aware cross-modal retrieval} task, introducing a dataset and some strong baselines that are designed to avoid a deep fine-tuning of the model on specific identity galleries. More in detail, we first introduced the \datasetacron{} dataset, obtained by replacing the faces in the COCO dataset with controlled, identity-specific entities from VGGFace2. We have shown that this task is particularly challenging for existing multimodal models like CLIP, which struggle to handle domain-specific entities and long-tail concepts such as unique identities not present in the training data.
As a second step, we also proposed \ourarchitectureacronym{}, a strong baseline that demonstrated significant improvements over standard CLIP models. In particular, we showed that while original CLIP can leverage biases related to names and ethnic groups, it is still limited in its ability to accurately retrieve images that both match the identity and the context described by the text query. Our experiments revealed that modifying CLIP augmenting the query with the face features of the person taken from an external gallery and incorporating a visual prompt tuning leads to notable enhancements in both identity and context retrieval.
As a future work, we plan to explore different pre-trained CLIP models and to augment \ourarchitectureacronym{} to actively weigh the importance of context and identity directly at inference time.
%Moreover, the results highlight that while CLIP can exploit name-based biases, these biases may introduce noise when paired with explicit identity representations, such as facial features. This underlines the complexity of multimodal identity-aware retrieval, where a balance between textual and visual information is critical. Our findings suggest that while models like Id-CLIP can substantially improve performance, there is still room for optimization, particularly in harmonizing text and image features when retrieving persons in specific contexts.

\begin{credits}
\subsubsection{\ackname} 
This work was partially funded by MUCES - a MUltimedia platform for Content Enrichment and Search in audiovisual archive project (CUP B53D23026090001), and by PNRR-M4C2 (PE00000013) "FAIR-Future Artificial Intelligence Research" - Spoke 1 "Human-centered AI", funded under the NextGeneration EU program.

\subsubsection{\discintname}
The authors have no competing interests to declare that are relevant to the content of this article.
\end{credits}

\bibliographystyle{splncs04}
\bibliography{bibliography,bibliography_itmatching}

% \section{note per il dataset}
% \subsection{coco}
% The annotations in this dataset along with this website belong to the COCO Consortium and are licensed under a\textbf{ Creative Commons Attribution 4.0 License}. \url{https://cocodataset.org/\#termsofuse} [vanno redistribute come CC]

% \textbf{The COCO Consortium does not own the copyright of the images}. Use of the images must abide by the \textbf{Flickr Terms of Use.} The users of the images accept full responsibility for the use of the dataset, including but not limited to the use of any copies of copyrighted images that they may create from the dataset.

% \subsection{Flickr Terms}
% Ci soon vari copyright sulle immagini..riuscimao a mappare quelle usate da noi?mm

% \subsection{Roop e AI Act}
% AI act: "Content that is either generated or modified with the help of AI - images, audio or video files (for example deepfakes) - need to be clearly labelled as AI generated so that users are aware when they come across such content."

% Roop: bIf using real faces, get consent and clearly label deepfakes when sharing. The developers aren't liable for user actions.
\end{document}